\documentclass[runningheads]{llncs}
\usepackage{graphicx}
\usepackage{microtype}
\usepackage{cite}
\usepackage{amsmath,amssymb,amsfonts}
\usepackage{algorithmic}
\usepackage{textcomp}
\usepackage{xcolor}
\usepackage{booktabs}
\usepackage{multirow}
\usepackage{url}

\begin{document}
\title{FineGen: A VLM-based Multi-Agent Framework for Fine-Grained Image-Text Dataset Construction}
%
%
\author{Chang Kong\inst{1,2}\orcidID{0000-0002-1923-9877} \and
Yuebing Li\inst{3}\orcidID{0009-0007-3816-3234} \and
Peng Mo\inst{3}\orcidID{0009-0007-4706-4975} \and
Haigang Zhang\inst{2}\orcidID{ORCID 0000-0002-8461-766X} \and
Qiuming Luo\inst{3}\orcidID{0000-0003-3622-5386}\thanks{Corresponding author}}
\authorrunning{C. Kong et al.}
%
\institute{Shenzhen Polytechnic University, Shenzhen, China 
\and
Institute of Applied Artificial Intelligence of the Guangdong-Hong Kong Macao Greater Bay Area, Shenzhen, China\\
\email{\{kongchang, zhanghg\}@szpu.edu.cn}\\
\and
Shenzhen University, China\\
\email{\{2410104002, mopeng2023\}@mails.szu.edu.cn, lqm@szu.edu.cn}}


%
\titlerunning{FineGen}
\maketitle              
\begin{abstract}
The scarcity of hard negative samples in current vision-language datasets significantly hinders fine-grained perception. To address this, we propose FineGen, a VLM-based Multi-Agent framework for automated dataset construction. By employing a collaborative ``Generation-Verification-Correction'' pipeline with a closed-loop feedback mechanism, FineGen ensures synthesized hard negatives are semantically valid yet strictly contradictory to visual content. Applying this to ImageNet, we construct FineGen-100K, a hierarchical dataset containing over 147,000 attribute-specific hard negatives with a rigorous 1:10 positive-to-negative ratio. Extensive evaluations confirm a 96.7\% attribute validity rate. Crucially, downstream validation on the FG-OVD benchmark shows that fine-tuning on FineGen-100K yields a substantial +14.4\% accuracy improvement on hard samples, significantly outperforming state-of-the-art methods.

\end{abstract}

\section{Introduction}

In recent years, Vision-Language Pre-training (VLP) models, such as CLIP~\cite{radford2021} and ALIGN~\cite{jia2021a}, have achieved remarkable success in various multimodal tasks by learning joint representations from large-scale image-text pairs. While these models demonstrate strong capabilities in capturing global semantics, they often struggle with \textit{fine-grained attribute perception}~\cite{thrush2022}\cite{yuksekgonul2023}\cite{bianchi2024}. For example, a model may correctly recognize the main object category in an image, but fail to distinguish subtle differences in color, material, texture, transparency, pattern, or object state. This limitation critically hinders the deployment of multimodal models in precision-sensitive applications, such as fine-grained product retrieval, open-vocabulary detection, and detailed image editing.

The primary bottleneck for advancing fine-grained perception lies in the inherent limitations of existing vision-language datasets. Mainstream datasets such as MS COCO~\cite{lin2014} and LAION~\cite{schuhmann2021} predominantly provide coarse-grained alignment, where textual descriptions focus on salient objects and overall scenes while frequently omitting detailed attributes. Moreover, standard contrastive learning paradigms typically rely on random batch sampling~\cite{chen2020}\cite{he2020}\cite{radford2021} to construct negative pairs. These coarse-grained negatives are often semantically distinct from the anchor image, allowing models to achieve high retrieval scores by leveraging global context without learning to discern specific attribute details. Crucially, there is a scarcity of minimal contrastive pairs~\cite{hsieh2023}\cite{bianchi2024}—sentences that share the same syntactic structure but differ only in specific attribute words—which serve as hard negatives to force models to align visual features with precise linguistic attributes.

Constructing a large-scale dataset with dense attribute annotations and hard negatives is a non-trivial task. Manual annotation is prohibitively expensive and difficult to scale, since annotators need to identify visible attributes, write grounded positive descriptions, design controlled attribute perturbations, and verify whether each negative description is truly inconsistent with the image. While recent approaches have explored using Large Language Models (LLMs) or Vision-Language Models (VLMs) to automate data generation~\cite{liu2023}\cite{chen2025i}, unconstrained generation often suffers from hallucinations~\cite{li2023g}\cite{guan2024a}, where the model generates plausible but visually incorrect details. More importantly, hard-negative construction introduces additional validity challenges: a generated negative may still be visually true for the image, become semantically implausible, or alter multiple attributes simultaneously, thereby weakening the intended fine-grained contrastive signal. Therefore, generating negative samples that are both `hard'' and `valid'' requires not only text generation, but also visual verification and targeted correction.

To address these challenges, we propose \textbf{FineGen}, a novel framework that leverages \textit{VLM-based Multi-Agent Collaboration} to automate the construction of fine-grained image-text datasets. Inspired by human collaborative workflows~\cite{madaan2023}\cite{li2023}, FineGen decomposes the complex data construction task into a modular pipeline involving \textit{Generation, Verification, and Correction} agents. Unlike monolithic generation approaches, FineGen introduces a closed-loop feedback mechanism: the Generation agent produces visually grounded positive descriptions and controlled hard-negative candidates; the Verification agent scrutinizes their visual-linguistic consistency and returns explicit feedback; and the Correction agent iteratively refines invalid or ambiguous samples according to the feedback. This Generate-Verify-Correct process effectively mitigates hallucinations and ensures the logical validity of synthesized hard negatives. Applying this framework to the ImageNet training set, we construct \textit{FineGen-100K}, a large-scale dataset featuring hierarchical fine-grained annotations, including global positive descriptions, hierarchical hard negatives, and atomic sub-text units.

The main contributions of this paper are summarized as follows:
\begin{itemize}
\item FineGen Framework: We propose a VLM-based multi-agent framework that automates fine-grained dataset construction. By decoupling the pipeline into Generation, Verification, and Correction roles, it achieves scalable and high-quality data synthesis without manual intervention.
\item Closed-Loop Verification: We introduce a ``Generate-Verify-Correct'' mechanism that enables agents to iteratively self-correct hallucinations and invalid hard negatives. This ensures that synthesized hard negatives are visually grounded minimal contrastive pairs rather than semantic noise.
\item FineGen-100K Dataset: We construct a large-scale hierarchical dataset featuring a dense 1:10 positive-to-negative ratio, with global positive descriptions, hierarchical hard negatives, and atomic sub-text units. Empirical results on the FG-OVD benchmark demonstrate state-of-the-art performance, surpassing previous methods by nearly 5\% on the challenging hard split.
\end{itemize}

\vspace{-5pt}
\section{Related Work}
\vspace{-5pt}
In this section, we review three lines of work related to FineGen: fine-grained vision-language benchmarks, automated data synthesis, and hard negative mining. These works motivate the need for training data that provides reliable attribute-level supervision rather than only coarse image-text alignment.

\vspace{-5pt}
\subsection{From Coarse-Grained to Fine-Grained Vision-Language Benchmarks}
\vspace{-5pt}

Standard vision-language datasets, such as MS COCO~\cite{lin2014} and LAION~\cite{schuhmann2021}, have driven the success of foundation models like CLIP~\cite{radford2021}. However, these datasets mainly focus on global semantic alignment, where captions describe salient objects and overall scenes while often omitting fine-grained attributes such as color, material, texture, and object state. As a result, models trained on them may achieve strong retrieval performance through global matching, but still fail to distinguish subtle compositional differences.

Recent benchmarks have made this limitation explicit. Winoground~\cite{thrush2022} tests whether models can distinguish image-caption pairs that share similar words but differ in compositional meaning. ARO~\cite{yuksekgonul2023} evaluates attribute binding and relational reasoning, revealing clear weaknesses in current VLP models. SugarCrepe~\cite{hsieh2023} and EqBen~\cite{wang2023c} introduce challenging negative captions to reduce shortcut-based evaluation. FG-OVD~\cite{bianchi2024}, which is closely related to our work, constructs fine-grained open-vocabulary detection benchmarks by rewriting structured object metadata from PACO into natural language. These benchmarks are valuable for diagnosing fine-grained perception failures, but they are mainly designed for evaluation or rely on existing structured annotations. In contrast, FineGen aims to generate large-scale fine-grained training data directly from raw images, bridging the gap between fine-grained evaluation and model training.

\vspace{-5pt}
\subsection{Automated Data Synthesis via Multi-Agent Collaboration}
\vspace{-5pt}

Synthetic data generation has become a practical way to reduce manual annotation cost. Early approaches use single VLMs, such as LLaVA~\cite{liu2023b} and ShareGPT4V~\cite{chen2025i}, to generate detailed image captions. Although scalable, single-pass generation often suffers from hallucinations, where the model describes objects or attributes that are not actually supported by the image.

To improve reliability, recent works have explored iterative or multi-agent collaboration. SynthSeg-Agents~\cite{wu2025} employs collaborative LLM agents to generate synthetic data for semantic segmentation, while AgentSGEN~\cite{xuan2025} uses an editor-evaluator loop for safety-critical scene generation. REVERSE~\cite{wuth2025} proposes a generate-but-verify paradigm to reduce hallucinations in VLM outputs. FineGen follows this collaborative direction, but focuses on fine-grained image-text dataset construction. Instead of using agents only for general caption refinement, FineGen explicitly separates Generation, Verification, and Correction. The Generator produces positive descriptions and attribute-level hard-negative candidates, the Verifier checks their visual-linguistic validity, and the Corrector revises invalid or ambiguous samples according to feedback. This closed-loop design is tailored to ensure that both positives and negatives are visually grounded.

\vspace{-5pt}
\subsection{Hard Negative Mining and Logical Validity}
\vspace{-5pt}

Contrastive learning relies heavily on the quality of negative samples. Standard approaches typically use batch negatives, where captions from other images are treated as negative samples. These negatives are often semantically distant from the anchor image and can be rejected through global context, providing limited pressure for the model to learn localized attributes.

Hard negative mining addresses this issue by constructing captions that are close to the positive description but factually inconsistent with the image~\cite{robinson2020a}. Methods like SugarCrepe~\cite{hsieh2023} generate hard negatives by replacing keywords, such as swapping `red'' with `blue'', using text-only LLMs. However, text-only modification can create false negatives. For example, changing `a happy dog'' to `a playful dog'' may still be visually true for the image, which weakens the contrastive signal. FG-CLIP~\cite{bianchi2024a} further suggests that fine-grained information may already exist in the representation space but requires more precise contrastive supervision to be effectively learned. FineGen addresses this problem by incorporating visual verification into negative construction. Each candidate hard negative is checked against the image before being accepted, so the final negatives are not only linguistically plausible but also visually contradictory to the specific image content.

\vspace{-5pt}
\section{Methodology}
\vspace{-5pt}

In this section, we present FineGen, a collaborative multi-agent framework designed to automate the construction of high-quality fine-grained image-text datasets. We first mathematically formulate the fine-grained alignment problem, then detail the system architecture, and finally elaborate on the prompt engineering strategies that govern agent behaviors.

\begin{figure*}[ht]
  \centering
  \includegraphics[width=1.0\textwidth]{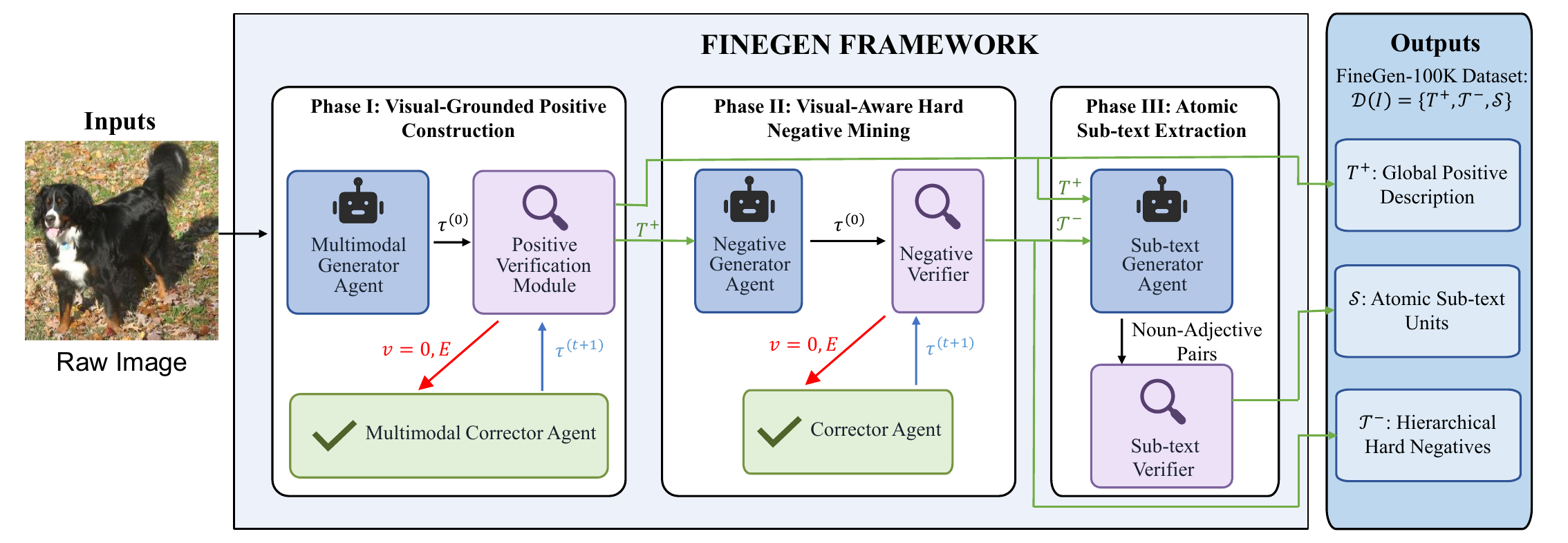}
  \vspace{-5pt}
  \caption{\textbf{Overview of the FineGen Framework.}
  The system automates fine-grained dataset construction via a \textbf{VLM-based Multi-Agent System}. The pipeline is horizontally decoupled into three main stages:
  \textbf{Phase I} synthesizes a visually-grounded global positive description ($T^+$);
  \textbf{Phase II} conducts visual-aware mining to generate a collection of hierarchical hard negatives ($\mathcal{T}^-$);
  and \textbf{Phase III} extracts atomic sub-text units ($\mathcal{S}$) for local grounding.
  Across all phases, FineGen employs a rigorous \textbf{Closed-Loop Collaboration Mechanism} involving three specialized roles:
  (1) \textbf{Generative Agents} produce initial candidates or attribute perturbations;
  (2) \textbf{Verification Agents} act as visual quality gates to detect hallucinations and logical inconsistencies;
  and (3) \textbf{Correction Agents} iteratively refine the text based on structured feedback.
  This Generate-Verify-Correct paradigm ensures that the final synthesized dataset $\mathcal{D}(I)=\{T^+,\mathcal{T}^-,\mathcal{S}\}$ is strictly grounded and semantically precise.}
  \label{fig:framework}
  \vspace{-15pt}
\end{figure*}

\vspace{-5pt}
\subsection{Problem Formulation}
\vspace{-5pt}

The core objective of this work is to construct a multi-granular dataset $\mathcal{D}$ that enables models to discern subtle attribute discrepancies. Formally, given an input image $I$, we aim to synthesize a triplet annotation $\mathcal{D}(I) = \{T^+, \mathcal{T}^-, \mathcal{S}\}$.

\textbf{Global Positive Description ($T^+$):} Let $\Omega$ be the space of all possible visual attributes (e.g., color, material, texture, state). $T^+$ is defined as a natural language sequence $w_{1:L}$ that maximizes the semantic alignment with $I$, subject to the constraint that it explicitly instantiates a subset of fine-grained attributes $A_{\text{sub}} \subset \Omega$ visible in $I$ (e.g., ``transparent glass'', ``metallic handle''). It is crucial to note that variables such as $T^+$ and $\mathcal{T}^-$ operate strictly in the discrete text/token space, rather than as continuous feature tensors in $\mathbb{R}^D$. Unlike generic captions, $T^+$ must be visually grounded at the attribute level.

\textbf{Hard Negative Samples ($\mathcal{T}^-$):} We define a hard negative as a text sequence that is syntactically isomorphic to $T^+$ but semantically contradictory to $I$ in a specific attribute dimension. Let $\phi(T, a \to a')$ be a perturbation function that replaces an attribute $a \in T$ with a decoy attribute $a'$. The set of hard negatives is constructed as $\mathcal{T}^- = \{\phi(T^+, a_i \to a'_i) \mid a_i \in A_{\text{sub}}, a'_i \neq a_i\}$. Crucially, a valid hard negative must satisfy two conditions: (1) \textit{Visual Contradiction}: The attribute $a'_i$ is visually absent in $I$ (e.g., ensuring the red car is not actually blue); (2) \textit{Logical Plausibility}: The combination of the object and $a'_i$ is semantically valid in the real world (e.g., excluding ``furry car'').

\textbf{Atomic Sub-text Units ($\mathcal{S}$):} To support region-level dense alignment, we decompose the global descriptions into a set of atomic units $\mathcal{S} = \{(u_k, y_k)\}_{k=1}^M$, where $u_k$ is a noun-adjective pair extracted from $T^+$ or $\mathcal{T}^-$, and $y_k \in \{0, 1\}$ is a binary label indicating whether $u_k$ is visually grounded in $I$. This formulation allows the dataset to provide supervision signals at both the global sentence level and the local phrase level.

\vspace{-5pt}
\subsection{The FineGen Framework}
\vspace{-5pt}

Constructing $\mathcal{D}$ via unconstrained generation often suffers from hallucinations and logic failures. To address this, FineGen adopts a VLM-based Multi-Agent System (MAS) architecture, prioritizing \textbf{modular decomposition} and \textbf{closed-loop verification}.

\vspace{-5pt}
\subsubsection{Architectural Design Philosophy}
As illustrated in Figure~\ref{fig:framework}, FineGen departs from monolithic ``black-box'' generation by decoupling the annotation process into specialized functional modules. We design the framework based on two core principles:
\textbf{Role Specialization}: We assign distinct cognitive roles—Generator ($G$), Verifier ($V$), and Corrector ($C$)—to separate agents. This separation of concerns prevents a single model from ``self-deception'' (i.e., verifying its own hallucinations as correct) and allows for the integration of specialized prompts for each task.
\textbf{Closed-Loop Quality Control}: A linear generation pipeline is prone to error accumulation. FineGen implements a recursive ``Generate-Verify-Correct'' feedback loop at every stage. This mechanism ensures that no data flows to the next stage until it passes a rigorous visual-semantic consistency check, effectively filtering out noise and guaranteeing that the final dataset is free from common synthetic artifacts.

\vspace{-5pt}
\subsubsection{System Components}
The framework consists of a central scheduler and three classes of VLM-driven agents:
\textbf{Generative Agents ($G$)}: These agents serve as the creative engine, responsible for translating visual signals into initial textual drafts ($T^+$) or performing linguistic perturbations to mine negatives ($\mathcal{T}^-$). They are optimized for fluency and attribute diversity.
\textbf{Verification Agents ($V$)}: These agents act as the ``judges'' equipped with strong visual-grounding capabilities. They independently assess the validity of the generated text against the raw image $I$, detecting hallucinations, missing attributes, or logical inconsistencies, and output structured error logs.
\textbf{Correction Agents ($C$)}: These agents function as the ``editors''. Instead of regenerating from scratch, they perform targeted rewriting by processing the error logs provided by the Verifiers, ensuring mathematical convergence towards high-quality samples.

\vspace{-5pt}
\subsection{Collaborative Dataset Construction}

\subsubsection{Phase I: Visual-Grounded Positive Construction}
The pipeline initiates with the generation of a visually accurate global description. A \textbf{Multimodal Generator Agent} ($G_{\text{pos}}$) analyzes the raw image $I$ to produce an initial candidate sequence $\tau^{(0)}$. Guided by specific prompts, the agent focuses on extracting fine-grained visual details such as texture, transparency, and object states:
$$\tau^{(0)} \leftarrow G_{\text{pos}}(I, \text{prompt})$$

To ensure reliability, the candidate text is immediately subjected to a \textbf{Positive Verification Module}. A VLM-based Verifier ($V_{\text{pos}}$) scrutinizes the alignment between $I$ and $\tau^{(t)}$, outputting a binary validity score $v \in \{0,1\}$ and a natural language Error Log $E$:
$$(v, E) \leftarrow V_{\text{pos}}(I, \tau^{(t)})$$

Upon detecting inconsistencies ($v=0$), the system triggers a \textbf{Correction Loop}. A Multimodal Corrector Agent ($C_{\text{pos}}$) receives the structured feedback $E$ (e.g., ``The object is metallic, not wooden'') and iteratively refines the text for the next state:
$$\tau^{(t+1)} \leftarrow C_{\text{pos}}(I, \tau^{(t)}, E)$$
This closed-loop process continues until the description passes verification ($v=1$), yielding the final sequence $T^+ \leftarrow \tau^{(t)}$, strictly grounded in visual evidence.

\vspace{-5pt}
\subsubsection{Phase II: Visual-Aware Hard Negative Mining}
Leveraging the validated positive description $T^+$, the system proceeds to construct hard negative samples $\mathcal{T}^-$. A \textbf{Negative Generator Agent} ($G_{\text{neg}}$) performs controlled perturbations on $T^+$ to produce an initial candidate $\tau^{(0)}$, targeting specific attribute dimensions (e.g., changing color from ``red'' to ``blue''):
$$\tau^{(0)} \leftarrow G_{\text{neg}}(I, T^+, a \to a')$$

Crucially, FineGen employs a \textbf{Visual-Logic Verification Mechanism}. The Negative Verifier ($V_{\text{neg}}$) performs a dual-check evaluation on the candidate against the image $I$:
$$(v, E) \leftarrow V_{\text{neg}}(I, \tau^{(t)})$$
The verification ensures: (1) \textit{Visual Contradiction}: confirming via VLM that the perturbed attribute is indeed false for $I$; and (2) \textit{Semantic Plausibility}: ensuring the modification yields a logically coherent sentence. If the negative is invalid or ambiguous ($v=0$), the Corrector Agent ($C_{\text{neg}}$) utilizes the explicit routing of $E$ to regenerate the sample:
$$\tau^{(t+1)} \leftarrow C_{\text{neg}}(I, \tau^{(t)}, E)$$
This state-transition feedback loop strictly ensures that $\mathcal{T}^-$ represents truly discriminative ``hard'' samples rather than random semantic noise.

\vspace{-5pt}
\subsubsection{Phase III: Atomic Sub-text Extraction}
To facilitate region-level alignment, the final phase decomposes the global descriptions into atomic units. A \textbf{Sub-text Generator} parses both $T^+$ and $\mathcal{T}^-$ to extract noun-adjective pairs (e.g., ``transparent glass'', ``opaque glass''). These units are rigorously filtered by a \textbf{Sub-text Verifier} to ensure linguistic completeness and correct binary labeling ($y = 1$ for positive pairs, $y = 0$ for negative pairs derived from perturbations). This hierarchical decomposition allows models to learn attribute binding at both the global sentence level and the local phrase level.

\vspace{-5pt}
\subsection{Constraint-based Prompt Engineering}
\vspace{-5pt}

The efficacy of multi-agent collaboration hinges on precise instruction following. Standard prompting often leads to uncontrollable outputs (e.g., altering sentence structures when generating negatives). To mitigate this, we employ a \textbf{Constraint-based In-Context Learning (ICL)} strategy, enforcing strict behavioral boundaries through three mechanisms:

\textbf{Persona-based Role Alignment}: To minimize the ambiguity of the LLM's latent space, each agent is initialized with a specific ``Persona''. For instance, the Positive Generator is prompted as an ``Expert Visual Describer focusing on physical attributes'', while the Negative Verifier is defined as a ``Strict Logic and Visual Inspector''. This persona injection stabilizes the stylistic consistency of the output, ensuring that descriptions remain objective and detailed rather than poetic or abstract.

\textbf{Hard Structural Constraints}: For tasks requiring syntactic precision---specifically Hard Negative Mining and Sub-text Extraction---we impose ``Hard Constraints'' within the prompts. For example, the Negative Generator is explicitly instructed: ``You must retain the original sentence structure and non-target words exactly. Only replace the target attribute word''. Similarly, the Sub-text Generator is constrained to output strictly formatted JSON objects containing only \{Noun, Adjective\} pairs. These structural priors significantly reduce the search space for the agents, preventing common errors such as sentence paraphrasing that would invalidate the minimal contrastive pair assumption.

\textbf{Feedback-Injection Mechanism}: A critical innovation in our prompt design is the dynamic integration of verification feedback. When a Corrector Agent is triggered, it does not merely receive the original task. Instead, the prompt is dynamically constructed to include: (1) The original image $I$; (2) The erroneous text $\tau^{(t)}$; and critically, (3) The explicit error reason $E$ provided by the Verifier (e.g., ``Error: The object 'wooden table' contradicts the visual evidence of 'glass table'''). This Feedback-Injection allows the LLM to perform ``Reasoning-Action'' (ReAct) updates, fixing the specific flaw without introducing new hallucinations, thereby accelerating the convergence of the closed-loop system.

\vspace{-5pt}
\section{Experiments}
\vspace{-5pt}

In this section, we conduct a comprehensive evaluation of the FineGen-100K dataset. Our evaluation consists of three parts: (1) statistical analysis of the dataset distribution; (2) intrinsic quality assessment via large-scale human evaluation; and (3) extrinsic validation by fine-tuning a CLIP-based model on downstream fine-grained perception benchmarks.

\begin{figure*}[ht]
  \centering
  \includegraphics[width=1.0\textwidth]{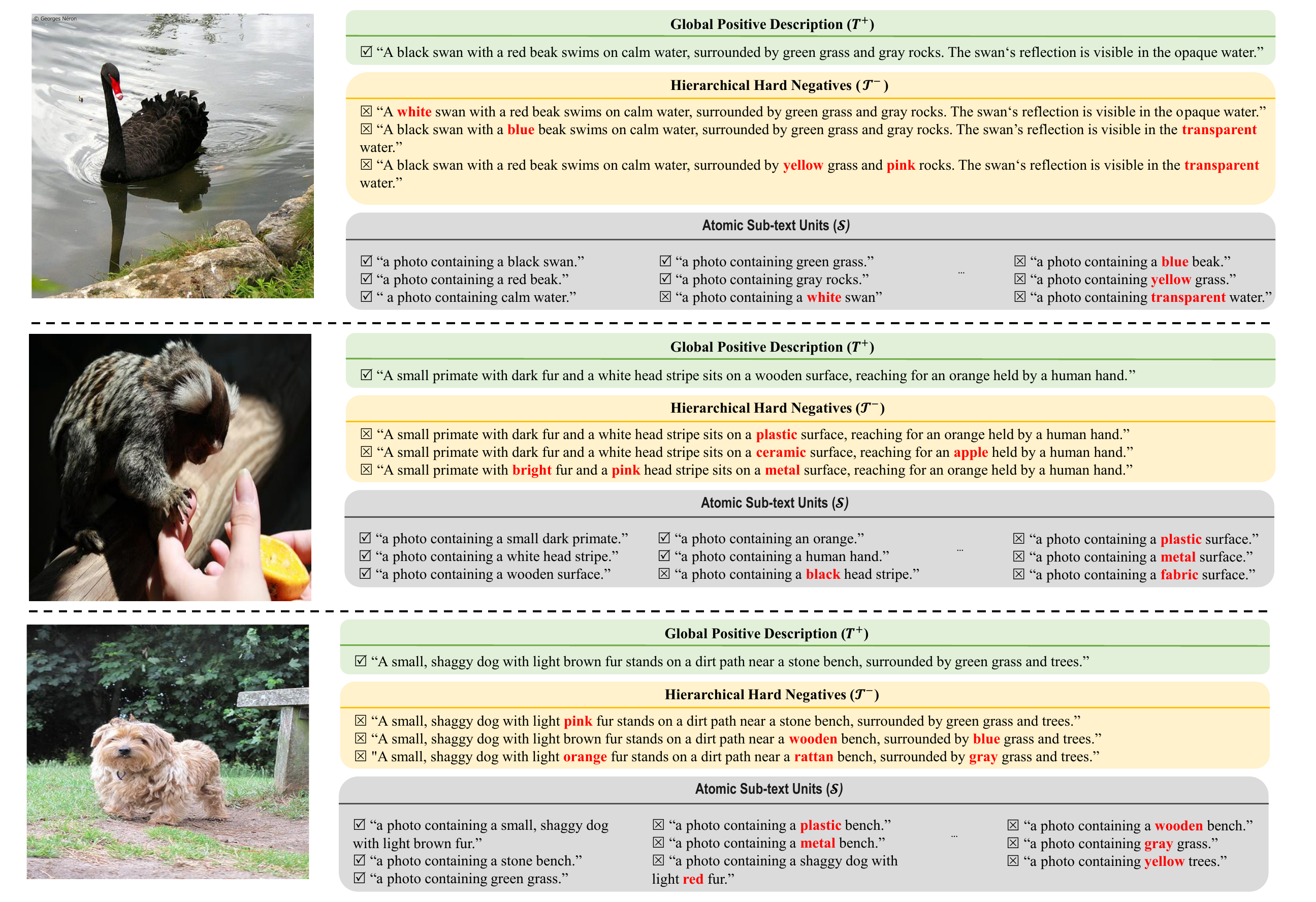}
  \vspace{-10pt}
  \caption{\textbf{Qualitative Visualization of FineGen-100K.} We visualize samples with three annotation levels: (1) \textbf{Positive Description}; (2) \textbf{Negative Descriptions} stratified by difficulty (Standard \textit{Easy} negatives involve multiple changes, while \textit{Hard} negatives involve single, subtle changes); and (3) \textbf{Atomic Sub-text Units} for local grounding.}
  \label{fig:visualization}
  \vspace{-10pt}
\end{figure*}

\vspace{-5pt}
\subsection{Dataset Statistics and Visualization}
\vspace{-5pt}

We constructed FineGen-100K by applying our proposed multi-agent framework to a subset of the ImageNet dataset. Unlike previous works that rely on MS COCO, utilizing ImageNet provides a diverse collection of object-centric images, which is particularly advantageous for isolating and analyzing fine-grained object attributes.

\vspace{-5pt}
\subsubsection{Quantitative Analysis}
As summarized in Table~\ref{tab:dataset_stats}, FineGen-100K comprises 9,997 unique images sampled from ImageNet. Each image is paired with one visually grounded positive description and an average of 10 attribute-specific hard negative samples, resulting in a total of 99,970 negative captions. This rigorous 1:10 positive-to-negative ratio provides dense contrastive signals, significantly exceeding the density of standard image-text datasets. In terms of attribute diversity, we explicitly modeled four key dimensions to ensure comprehensive coverage. The distribution of attribute modifications in the negative samples is as follows: Color (66.48\%), Material (24.04\%), Pattern (6.59\%), and Transparency (2.89\%). This distribution reflects the visual complexity of real-world objects and ensures the model learns to discern both salient features (e.g., color) and subtle material properties (e.g., transparency).

\begin{table}[t]
\centering
\caption{Statistics of FineGen-100K Dataset.}
\vspace{-5pt}
\label{tab:dataset_stats}
\setlength{\tabcolsep}{10pt}
\begin{tabular}{lrr}
\toprule
\textbf{Metric} & \textbf{Count} & \textbf{Pct.} \\
\midrule
\multicolumn{3}{l}{\textit{Dataset Scale}} \\
Total Images (ImageNet) & 9,997 & -- \\
Positive Captions & 9,997 & -- \\
Hard Negative Captions & \textbf{99,970} & -- \\
Pos/Neg Ratio & \textbf{1:10} & -- \\
\midrule
\multicolumn{3}{l}{\textit{Attribute Breakdown}} \\
Color & 97,984 & 66.48\% \\
Material & 35,424 & 24.04\% \\
Pattern & 9,709 & 6.59\% \\
Transparency & 4,261 & 2.89\% \\
\midrule
\textbf{Total Edits} & \textbf{147,381} & \textbf{100.0\%} \\
\bottomrule
\end{tabular}
\vspace{-5pt}
\end{table}

\vspace{-5pt}
\subsubsection{Qualitative Visualization}
To illustrate the multi-granular structure of our data, Figure~\ref{fig:visualization} presents representative samples. Each sample is annotated with three levels of information:
\begin{enumerate}
    \item \textbf{Global Positive Description}: A sentence accurately describing the central object and its fine-grained attributes (e.g., ``A black swan with a red beak...'').
    \item \textbf{Hierarchical Hard Negatives}: From the pool of 10 generated negatives, we display three representative examples corresponding to \textit{Easy}, \textit{Medium}, and \textit{Hard} difficulty levels. These levels are defined by the granularity of attribute perturbation: `Hard' negatives typically involve a single, subtle attribute change (e.g., only changing opaque to transparent), whereas `Easy' negatives may involve multiple simultaneous modifications, making them easier to distinguish.
    \item \textbf{Atomic Sub-text Units}: A collection of noun-adjective pairs extracted from the descriptions. Each unit is explicitly labeled as visually consistent (Positive) or hallucinated (Negative), supporting region-level attribute learning.
\end{enumerate}
This hierarchical visualization demonstrates that FineGen-100K not only captures attribute discrepancies but also organizes them by difficulty, equipping models with the ability to distinguish fine-grained details in an open-world setting.

\vspace{-5pt}
\subsection{Human Evaluation of Data Quality}
\vspace{-5pt}

To validate the reliability of our automated pipeline, we conducted a rigorous human inspection on a randomly sampled subset of FineGen-100K. The evaluation set comprised a total of 11,423 annotation units, consisting of 5,500 global descriptions (stratified across positive captions and varying difficulty levels of negative samples) and 5,923 atomic sub-text units.

Human experts were instructed to identify ``unreasonable replacements,'' which include: (1) Hallucinations in positive descriptions (describing non-existent attributes); (2) Invalid Perturbations in negative samples (e.g., attributes that remain visually true after modification or result in semantic nonsense); and (3) Mislabeled sub-text units.

\textbf{Results}
Across the entire evaluation set, we identified only 377 instances of erroneous or unreasonable replacements. This corresponds to a remarkably low error rate of 3.3\% (i.e., an overall accuracy of 96.7\%). This high fidelity confirms that our multi-agent closed-loop mechanism effectively filters out the noise typically associated with VLM-generated data, ensuring that the dataset provides precise and reliable supervision signals.

\vspace{-5pt}
\subsection{Downstream Task Evaluation}
\vspace{-5pt}

To validate the practical utility of FineGen-100K, we fine-tuned a standard CLIP (ViT-B/16) model using the SubCLIP strategy and evaluated it on the FG-OVD Benchmark. We compared our method against zero-shot CLIP and state-of-the-art fine-grained alignment methods.

\textbf{Results and In-depth Attribution.} Table~\ref{tab:downstream_results} presents the comparison results. Note that SubCLIP denotes the model fine-tuned on our FineGen-100K dataset, and all reported methods utilize the same ViT-B/16 backbone for fair comparison. Our method demonstrates superior performance, particularly on the most challenging Hard split, reaching a Top-1 Accuracy of 34.53\%. This represents a substantial +14.4\% gain over the CLIP baseline and outperforms the previous best method (Fine-CLIP) by nearly 5\%. Consistent improvements are also observed on the Medium and Easy splits. 

To conceptually understand this significant +14.4\% performance gain, we analyze the optimization dynamics of contrastive learning. Standard coarse negatives allow vision-language models to rely on \textit{global semantic shortcuts}. This permits trivial ``bag-of-words'' matching, which subsequently causes vanishing gradients for localized, fine-grained features. Conversely, FineGen-100K provides syntactically isomorphic negatives that differ by only a single specific attribute. These highly controlled samples yield \textit{hard contrastive gradients} during training. This mechanism explicitly forces the model to abandon global approximations and strictly converge on localized visual features. Consequently, these results strongly demonstrate that the dense, high-quality hard negatives in FineGen-100K effectively empower VLMs to transcend coarse semantic matching and master fine-grained attribute discrimination.

\begin{table}[t]
\centering
\caption{Evaluation on FG-OVD Benchmark (Box-Cropped Image-Level Matching). \textbf{MR denotes Miss Rate (lower is better).}}
\vspace{-5pt}
\label{tab:downstream_results}
\resizebox{\columnwidth}{!}{%
\begin{tabular}{lcccccccc}
\toprule
\multicolumn{1}{c}{\multirow{2}{*}{\textbf{Method}}} & \multicolumn{2}{c}{\textbf{Hard}} & \multicolumn{2}{c}{\textbf{Medium}} & \multicolumn{2}{c}{\textbf{Easy}} & \multicolumn{2}{c}{\textbf{Trivial}} \\
\cmidrule(lr){2-3} \cmidrule(lr){4-5} \cmidrule(lr){6-7} \cmidrule(lr){8-9}
 & \textbf{Acc.} $\uparrow$ & \textbf{MR} $\downarrow$ & \textbf{Acc.} $\uparrow$ & \textbf{MR} $\downarrow$ & \textbf{Acc.} $\uparrow$ & \textbf{MR} $\downarrow$ & \textbf{Acc.} $\uparrow$ & \textbf{MR} $\downarrow$ \\
\midrule
CLIP\cite{radford2021}      & 20.1 & 3.32 & 43.1 & 2.52 & 45.6 & 2.46 & 75.8 & 1.63 \\
EVA-CLIP\cite{sun2023}  & 18.0 & 3.69 & 38.4 & 2.95 & 38.4 & 3.06 & 82.8 & 1.38 \\
Long-CLIP\cite{zhang2024} & 16.3 & 3.81 & 31.7 & 3.19 & 36.0 & 3.15 & 77.1 & 1.55 \\
Fine-CLIP\cite{jing2024} & 29.6 & 2.88 & \textbf{52.8} & 2.21 & 56.4 & 2.15 & 86.0 & \textbf{1.29} \\
FG-CLIP\cite{bianchi2024a}   & 26.7 & 2.92 & 51.2 & 2.21 & 56.6 & 2.16 & \textbf{86.5} & \textbf{1.29} \\
\midrule
\textbf{SubCLIP}\cite{luo2025} & \textbf{34.53} & \textbf{2.54} & 51.4 & \textbf{2.20} & \textbf{57.6} & \textbf{2.13} & 76.1 & 1.61 \\
\bottomrule
\end{tabular}%
}
\vspace{-5pt}
\end{table}

\vspace{-5pt}
\subsection{Discussion and Limitations}
\vspace{-5pt}

While FineGen demonstrates robust performance and high logical validity, we acknowledge a potential limitation regarding \textit{Verifier Bias}. Because the Verification Agents within our closed-loop system are driven by specific VLMs, the generated dataset might inadvertently inherit the latent blind spots or biases inherent to those foundational models. If the Verifier consistently misjudges a specific rare attribute, this systematic error could propagate into the final dataset. To mitigate this in future iterations, we plan to implement an ensemble strategy utilizing heterogeneous VLMs (e.g., cross-checking between Qwen-VL and GPT-4o) for mutual verification. Nevertheless, the substantial +14.4\% performance gain on the highly diverse, real-world FG-OVD benchmark indicates that any inherited bias does not currently degrade downstream generalizability.

\vspace{-5pt}
\section{Conclusion}
\vspace{-5pt}

In this paper, we address the scarcity of fine-grained vision-language data by proposing FineGen, a VLM-based multi-agent framework that automates the construction of high-quality datasets from ImageNet. By implementing a novel closed-loop feedback mechanism, FineGen effectively mitigates hallucinations and ensures the logical validity of synthesized hard negatives. The resulting FineGen-100K dataset features a hierarchical structure with dense attribute-specific annotations. Extensive evaluations—yielding a 96.7\% attribute validity rate and a 14.4\% performance gain on the FG-OVD hard split—conclusively demonstrate that FineGen-100K effectively equips VLMs with precise fine-grained discriminative capabilities. Future work will extend this paradigm to complex modalities like video understanding.

%
%
%
%

\bibliographystyle{splncs04}
\bibliography{refs}
\appendix
\vspace{-5pt}
\section{Appendix: Cost-Benefit Analysis}
\vspace{-5pt}

In our framework, constructing the FineGen-100K dataset requires automated VLM inference for the Generation, Verification, and Correction agents. Processing the entire training set (147,381 localized editing tasks across approximately 10,000 images) via the Qwen3.5-Plus API incurred a total expenditure of approximately \$100. This results in an average cost of less than \$0.01 per image. 

In comparison, traditional manual expert annotation—required to achieve the same density of fine-grained attribute-specific hard negatives—is conservatively estimated to cost between \$0.50 and \$1.00 per image, necessitating 2-3 minutes of human labor per triplet. Thus, FineGen achieves a greater than 50$\times$ reduction in financial expenditure while simultaneously accelerating the data synthesis cycle, thereby proving the framework's scalability for practical, large-scale VLM training applications.
\end{document}